\documentclass[
]{ceurart}

\usepackage{listings}
\usepackage{subcaption}
\usepackage{graphicx}
\usepackage{algorithm2e}
\usepackage{arydshln}
\begin{document}

\copyrightyear{2023}
\copyrightclause{Copyright for this paper by its authors.
  Use permitted under Creative Commons License Attribution 4.0
  International (CC BY 4.0).}

\conference{CLEF 2023: Conference and Labs of the Evaluation Forum, September 18–21, 2023, Thessaloniki, Greece}

\title{UIT-Saviors at MEDVQA-GI 2023: Improving Multimodal Learning with Image Enhancement for Gastrointestinal Visual Question Answering}



\author[1,2]{Triet M. Thai}[%
email=19522397@gm.uit.edu.vn,
]

\address[1]{Faculty of Information Science and Engineering, Unviversity of Information Technology, Ho Chi Minh City, Vietnam}
\address[2]{Vietnam National University, Ho Chi Minh City, Vietnam}

\author[1,2]{Anh T. Vo}[%
email=19521226@gm.uit.edu.vn,
]

\author[1,2]{Hao K. Tieu}[%
email=19521480@gm.uit.edu.vn,
]

\author[1,2]{Linh N.P. Bui}[%
email=20521527@gm.uit.edu.vn,]

\author[1,2]{Thien T.B. Nguyen}[%
email=thienntb@uit.edu.vn,
]
\cormark[1]
\cortext[1]{Corresponding author.}

\begin{abstract}
In recent years, artificial intelligence has played an important role in medicine and disease diagnosis, with many applications to be mentioned, one of which is Medical Visual Question Answering (MedVQA). By combining computer vision and natural language processing, MedVQA systems can assist experts in extracting relevant information from medical image based on a given question and providing precise diagnostic answers. The ImageCLEFmed-MEDVQA-GI-2023 challenge carried out a visual question answering (VQA) task in the gastrointestinal domain, which includes gastroscopy and colonoscopy images. Our team approached Task 1 - Visual Question Answering of the challenge by proposing a multimodal learning method with image enhancement to improve the VQA performance on gastrointestinal images. The multimodal architecture is set up with a BERT encoder and different pre-trained vision models based on convolutional neural network (CNN) and Transformer architecture for features extraction from question and endoscopy image. The result of this study highlights the dominance of Transformer-based vision models over the CNNs and demonstrates the effectiveness of the image enhancement process, with six out of the eight vision models achieving better F1-Score. Our best method, which takes advantages of BERT+BEiT fusion and image enhancement, achieves up to 87.25\% accuracy and 91.85\% F1-Score on the development test set, while also producing good result on the private test set with accuracy of 82.01\%.
\end{abstract}


\maketitle

\section{Introduction}
The digestive system is one of the most complex and essential systems in the human body, consisting of various organs such as the mouth, stomach, intestines, and rectum. From the process of digestion in the stomach to the absorption of nutrients in the small and large intestines, and finally the elimination of waste through the rectum, the entire process involves the interaction and coordination of each organ to ensure the supply of nutrients and energy to the body. Any issues that occur in any part of the digestive system can directly impact the entire gastrointestinal tract, such as inflammation of the intestines, digestive cancers, and diseases of the stomach and colon, especially colorectal diseases, which remain a significant concern for the healthcare community. According to estimates from the American Cancer Society\footnote{\href{https://www.cancer.org/cancer/types/colon-rectal-cancer/about/new-research.html}{https://www.cancer.org/cancer/types/colon-rectal-cancer/about/key-statistics.html}. }, colorectal cancer ranks as the third leading cause of cancer-related deaths for both men and women in the United States. The projected numbers for colorectal cancer cases in the year 2023 are 106,970 new cases of colon cancer and 46,050 new cases of rectal cancer, with an estimated 52,550 deaths.
However, it is important to note that the mortality rate from colorectal cancer has decreased over the past decade due to advancements in scientific and technological research. Screening techniques allow for the detection of abnormalities in the colon and rectum to be removed before they develop into cancer.

Clinical imaging techniques such as X-rays, computed tomography (CT), or ultrasound are often not highly effective in diagnosing pathological conditions in the colon. Therefore, colonoscopy remains the primary technique used for detection, screening, and treatment of gastrointestinal diseases. This method involves using a flexible endoscope, which is inserted through the anus and advanced into the colon. The real-time images of the colon obtained from the endoscopic device are displayed on a monitor, allowing the physician to observe and evaluate any abnormalities in the intestinal tract, the condition of the mucosal lining, and other structures within the colon.

Colonoscopy is considered the gold-standard screening procedure for examining and treating colorectal diseases. The endoscopic images contain a wealth of important information about the patient's condition. However, the effectiveness of the colonoscopy process can vary depending on the skills of the performer and the complexity of the endoscopic image analysis, which requires specialized knowledge and manual interpretation \cite{Rex2015}. To improve the performance of colonoscopy in accurately detecting and classifying lesions, decision support systems aided by artificial intelligence (AI) are being rapidly developed. Among them, Visual Question Answering (VQA) is one of the most prominent techniques. Combining computer vision and natural language processing, VQA assists in extracting information from images, identifying abnormalities, and providing accurate answers to specific diagnostic questions. By integrating information from images and questions, VQA enhances the accuracy of lesion detection and classification, improves communication between users and images, and helps guide appropriate treatment strategies.

To successfully deploy VQA in the healthcare domain, in addition to algorithmic integration, a sufficiently large and diverse training dataset is required. Our research team participated in the VQA task of the ImageCLEFmed Medical Visual Question Answering on Gastrointestinal Image (MEDVQA-GI) \cite{ImageCLEFmedicalCaptionOverview2023} competition at ImageCLEF2023\cite{ImageCLEF2023}. The contribution of the paper focused on performing the VQA task with a new dataset from ImageCLEFmed MEDVQA-GI. Specifically, we employed a multimodal approach for the VQA task (Task~1), combining information from two primary data sources: endoscopic images and textual questions. To achieve a good performance on the VQA task with the provided dataset, we first performed an efficient image preprocessing steps, which involved specular highlights inpainting, noise, and black mask removal to enhance the image quality. Subsequently, we conducted experiments and compared the performance of various image feature extraction models based on CNN and Transformer using both raw and enhanced image data. The final results, with accuracy up to 87.25\%  on the development test set and 82.01\% on the private test set, demonstrate the potential of the proposed method in improving the performance of VQA systems in the field of gastrointestinal endoscopy imaging in general and colonoscopy in particular.

\section{Background and Related Works}

\subsection{Colonoscopy Image Analysis}
With the advancement of modern advanced technology, AI has made significant contributions to the field of healthcare, specifically in the progress of the colonoscopy examination process. Currently, two potential approaches with AI being utilized for colonoscopy image analysis, including Computer-Aided Detection (CAD) and Deep Learning (DL) systems. In the CAD approach, the system utilizes image processing algorithms to improve the performance of endoscopic procedures, enabling physicians to easily detect lesions in hard-to-identify locations and reduce the chances of misdiagnosis
\cite{hassan2021performance}. On the other hand, the DL-based system employs a deep learning model trained on specific datasets, which enhances the accuracy of lesion detection compared to the CAD-based system  \cite{lee2020real}. However, developing algorithms for automatic analysis and anomaly detection in endoscopic images requires preliminary image preprocessing to address various factors, such as specular highlights, interlacing or artefacts that impact the system's performance \cite{Soto17}.
 
\subsection{Preprocessing Methods for Colonoscopy Images}
In reality, the quality of endoscopy images depends on various factors such as the skill of the performing physician, limitations of the equipment, and certain environmental conditions. Some common difficulties in processing endoscopy images include black masks, ghost colors, interlacing, specular highlights, and uneven lighting  \cite{soeder2022high}. 
Black masks are the occurrence of a black border around the edges of the image due to the use of lenses in the endoscopy system that have a black frame surrounding the edges. This frame can hinder the development of algorithms. To address this issue, techniques such as restoration, thresholding, cropping, or inpainting are necessary. Specular highlights, which are bright spots reflected from tumors or polyps captured by the camera, can disrupt the algorithms. Therefore, to remove them, we can employ detection or inpainting methods. Additionally, for issues like interlacing, ghost colors, and uneven lighting, segmentation methods can be applied to achieve optimal results  \cite{Soto17} \cite{6611256} \cite{Arnold2010AutomaticSA}.
Overall, preprocessing steps play a crucial role in mitigating the challenges commonly encountered with colonoscopy images. The mentioned techniques will help improve the overall quality of the images, thereby enhancing the performance of analysis and diagnosis.

\subsection{Medical Visual Question Answering} 
Medical visual question answering (MedVQA) is an important field in medical AI that combines VQA challenges with healthcare applications. By integrating medical images and clinically relevant questions, MedVQA systems aim to provide plausible and convincing answers. While VQA has been extensively studied in general domains, MedVQA presents unique opportunities for exploration.
Currently, there are 8 publicly available MedVQA datasets, including VQA-MED-2018~\cite{hasan2018overview}, VQA-RAD \cite{lau2018dataset}, VQA-MED-2019 \cite{abacha2019vqa}, RadVisDial \cite{kovaleva2020towards}, PathVQA \cite{he2020pathvqa}, VQA-MED-2020 \cite{ben2021overview}, SLAKE \cite{liu2021slake}, and VQA-MED-2021 \cite{ben2021overview}. These datasets serve as valuable resources for advancing MedVQA research.

The basic framework of MedVQA systems typically contains an image encoder, a question encoder, a fusion algorithm, and an answering component. Other frameworks may exclude the question encoder when the question is simple. Common choices for image encoder are ResNet \cite{he2016deep} and VGGNet \cite{simonyan2014very} that are pre-trained on ImageNet dataset \cite{russakovsky2015imagenet}. For language encoders, Transformer-based architectures such as BERT \cite{kenton2019bert} or BioBERT \cite{lee2020biobert} are commonly applied because of their proven advantages, besides the Recurrent Neural Networks (LSTM \cite{hochreiter1997long}, Bi-LSTM \cite{schuster1997bidirectional}, GRU \cite{cho2014learning}). The fusion stage, the core component of VQA methods, has typical fusion algorithms, including the attention mechanism and the pooling module. Common attention mechanisms are the Stacked Attention Networks (SAN) \cite{yang2016stacked}, the Bilinear Attention Networks (BAN) \cite{sacramento2018dendritic}, or the Hierarchical Question-Image Co-Attention (HieCoAtt) \cite{lu2016hierarchical}. Most multimodal pooling practices are concatenation, sum, and element-wise product. The attention mechanism can aggregate with the pooling module.
The answering component has two modes of output depending on the properties of the answer. The classification mode is used if the answer is brief and limited to one or two words. Otherwise, if the response is in free-form format, the generation modules such as LSTM or GRU are taken into account. There are additional techniques to the basic concept, for instance, Sub-task strategy, Global Average Pooling \cite{lin2013network}, Embedding-based Topic Model, Question-Conditioned Reasoning, and Image Size Encoder.
\section{Task and Dataset Descriptions}

\subsection{Task Descriptions}

Identifying lesions in endoscopy images is currently one of the most popular applications of artificial intelligence in the medical field. For the task at ImageCLEFmed-MEDVQA-GI-2023 \cite{ImageCLEFmedicalCaptionOverview2023}, the main focus will be on VQA and visual question generation (VQG). The main goal is to provide support to healthcare experts in diagnosis by combining image and text data for analysis. The task consists of three sub-tasks:
\begin{enumerate}
    \item \textbf{VQA (Visual Question Answering):} For the visual question answering part, participants are required to generate a textual answer to a given textual question-image pair. This task involves combining endoscopy images from the dataset with textual answers to respond to questions.
    \item \textbf{VQG (Visual Question Generation):} This is the reverse task of VQA, where participants need to generate textual questions based on given textual answers and image pairs.
    \item \textbf{VLQA (Visual Location Question Answering):} Participants are provided with an image and a question, and they are required to provide an answer by providing a segmentation mask for the image.
\end{enumerate}
In this study, our team only focuses on the VQA task (Task 1) for the provided endoscopy image dataset. In general, we receive a textual question along with the corresponding image, and the main task is to generate accurate and appropriate answers based on information from both sources. For example, for an image containing a colon polyp with the following question, "Where in the image is the polyp located?", the proposed VQA system should return answer giving a textual description of where in the image the polyp is located, like upper-left or in the center of the image.

\subsection{Dataset Information}
\begin{table*}[t]
\caption{Questions and sample answers from ImageCLEFmed-MEDVQA-GI-2023 dataset}
\label{tab:ques}
\resizebox{\textwidth}{!}{%
\begin{tabular}{rll}\toprule
\textbf{ID} & \textbf{Questions} & \textbf{Sample Answers}\\\midrule
0                               & What type of procedure is the image taken from? & "Colonoscopy", "Gastroscopy"\\
1                               & Have all polyps been removed?& "Yes", "No", "Not relevant"                 \\
2                               & Is this finding easy to detect? & "Yes", "No", "Not relevant"                             \\
3                               & Is there a green/black box artifact? & "Yes", "No"        \\
4                               & Is there text? & "Yes", "No"                                \\
5                               & What color is the abnormality? & "Red", "Pink", "White, Yellow", ...                 \\
6                               & What color is the anatomical landmark? &  "Red", "Red, White", "Pink, Red, grey", ...         \\
7                               & How many findings are present? & "0", "1", "2", "3", "4", "5", ...                                     \\
8                               & How many polyps are in the image? & "0", "1", "2", "3", "4", "5", ...               \\
 9                               & How many instruments are in the image? & "0", "1", "2", "3"    \\
 10                              & Where in the image is the abnormality? & "Center", "Lower-left", "Lower-right, Center-right", ... \\
 11                              & Where in the image is the instrument? & "Center", "Lower-left", "Lower-right, Center-right", ... \\
 12                              & Are there any abnormalities in the image? & "No", "Polyp", "Ulcerative colitis", "Oesophagitis", ... \\
  13                              & Are there any anatomical landmarks in the image? & "No", "Z-line" "Cecum", "Ileum", "Pylorus", "Not relevant"\\
  14                              & Are there any instruments in the image? & "No", "Tube", "Biopsy forceps", "Metal clip", 'Polyp snare, Tube',  ...    \\
  15                              & Where in the image is the anatomical landmark? & "Center", "Lower-left", "Lower-right, Center-right", ...   \\
  16                              & What is the size of the polyp? & "< 5mm", "5-10mm", "11-20mm", ">20mm", "Not relevant", ...    \\
  17                              & What type of polyp is present? & "Paris ip", "Paris iia", "Paris is", "Paris is, Paris iia", ...    \\\bottomrule
 
\end{tabular}}
\end{table*}

\begin{figure}[t]
\centering
\begin{subfigure}[t]{.235\linewidth}
\centering
\frame{\includegraphics[width=0.95\linewidth]{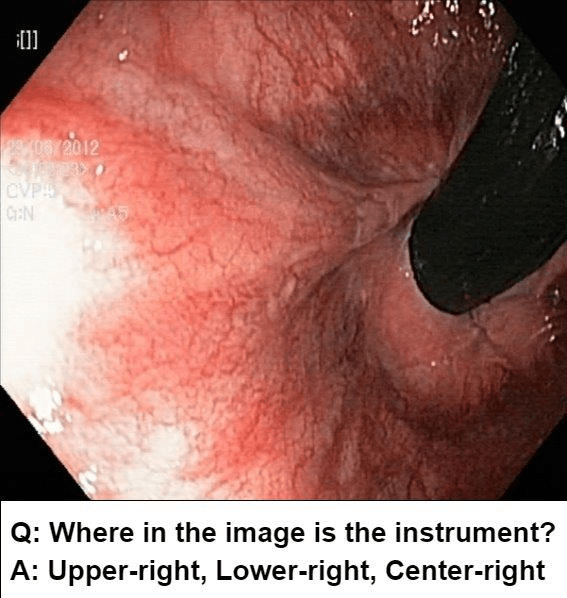}}
\caption{No abnormality}
\label{sfig:ab1}
\end{subfigure}%
\begin{subfigure}[t]{.235\linewidth}
\centering
\frame{\includegraphics[width=.95\linewidth]{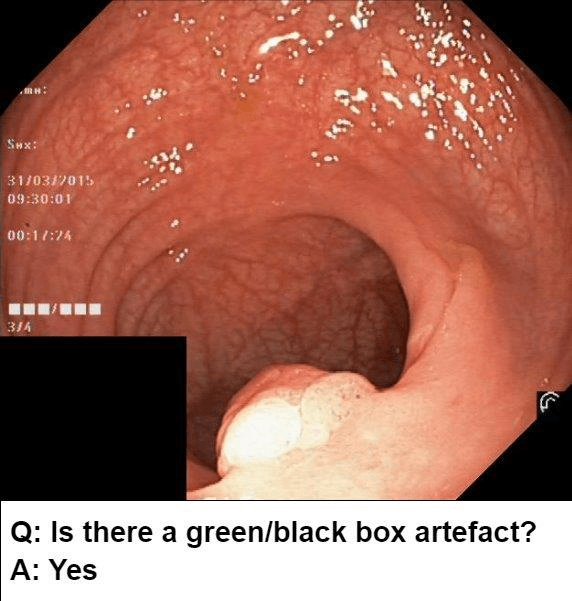}}
\caption{Polyp}
\label{sfig:ab2}
\end{subfigure}
\begin{subfigure}[t]{.257\linewidth}
\centering
\frame{\includegraphics[width=0.95\linewidth]{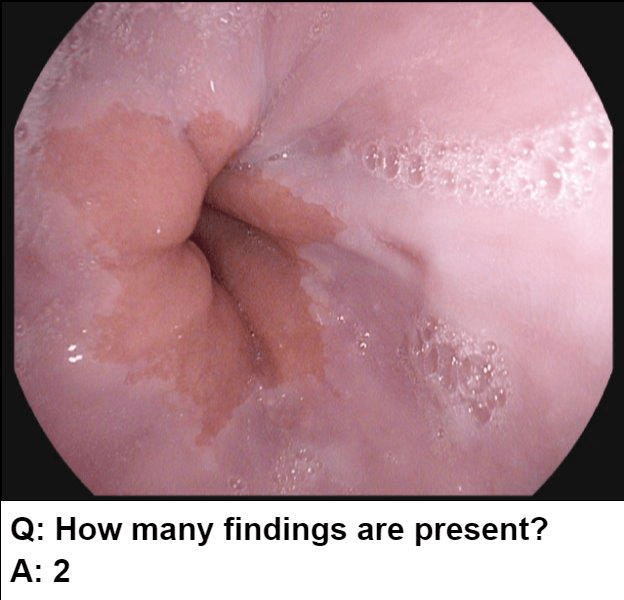}}
\caption{Oesophagitis}
\label{sfig:ab3}
\end{subfigure}%
\begin{subfigure}[t]{.257\linewidth}
\centering
\frame{\includegraphics[width=0.95\linewidth]{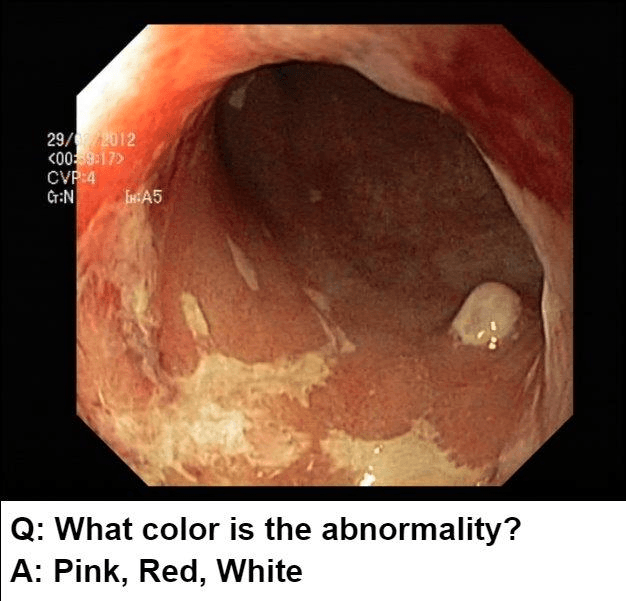}}
\caption{Ulcerative Colitis}
\label{sfig:ab4}
\end{subfigure}%
\caption{Illustrations of question-answer pairs along with common abnormalities in gastrointestinal image from ImageCLEFmed-MEDVQA-GI-2023 dataset}
\label{fig:ab}
\end{figure}

The new dataset released for the ImageCLEFmed-MEDVQA-GI-2023 challenge is based on the HyperKvasir dataset \cite{Borgli2020}, the largest gastrointestinal collections with more than 100,000 images, with the additional question-and-answer ground truth developed by medical collaborators. The development set and test set include a total of 3949 images from different procedures such as gastroscopy and colonoscopy,  spanning the entire gastrointestinal tract, from mouth to anus. Each image has a total of 18 questions about abnormalities, surgical instruments, normal findings and other artefacts, with multiple answers possible for each, as shown in Table \ref{tab:ques}. Not all questions will be relevant to the provided image, and the VQA system should be able to handle cases where there is no correct answer. Figure \ref{fig:ab} depicts several examples of question-answer pairs on common abnormalities in gastrointestinal tract, such as Colon Polyps, Oesophagitis, and Ulcerative Colitis. As shown in Figure \ref{sfig:ab4}, there are three possible answers to the question "What color is the abnormality?": "Pink," "Red," and "White", and a typical VQA system should be able to identify all three colors.  In general, the image may contains a variety of noise and components that locates across abnormalities, such as highlight spots or instruments, which pose a significant challenge in developing efficient VQA systems for gastrointestinal domain.

\section{The Proposed Approach}
\begin{figure}[t]
\centering
\includegraphics[width=\linewidth]{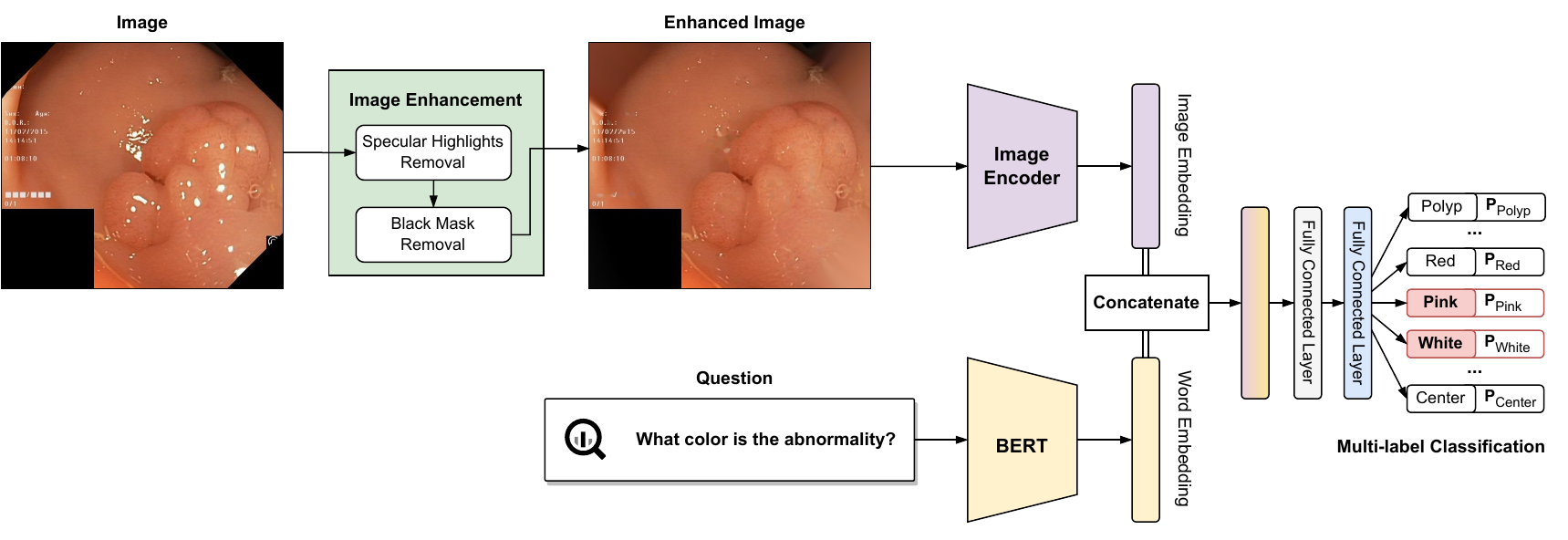}

\caption{An overview of the multimodal architecture with image enhancement for VQA challenge}
\label{fig:method}
\end{figure}

The method used in this study is based on a standard framework that is commonly used to tackle general VQA problems. Figure \ref{fig:method} depicts an overview of the proposed method for ImageCLEFmed-MEDVQA-GI-2023 dataset. In general, the VQA architecture employs powerful pre-trained models to extract visual and textual features from image-question pairs, which are then combined into a joint embedding using a fusion algorithm and passed to a classifier module to generate the appropriate answer. To improve the quality of the region of interest and achieve better VQA performance, the original image is passed through a series of enhancement procedures before being fed into the image encoder for features extraction. 

\subsection{Image Enhancement}

The purpose of the image pre-processing and enhancement steps is to remove noise and artifacts, which are frequently caused by the equipment used in diagnostic or environmental difficulties. Some of the major problems to be mentioned are black mask, specular highlights, interlacing or uneven lighting. The impact of these elements, such as black mask and specular highlights, is significant since they, like the polyp, create valley information and affect the performance of polyp localization, causing the VQA system to generate incorrect answers. 

This study employs pre-processing and enhancing methods to cope with specular highlights and black mask in colonoscopy image, which are prevalent artifacts in the dataset provided. The desired outcome is an enhanced image with no specular reflection or black frame while retaining the visual features of the region of interest.

\subsubsection{Specular Highlights Removal}
\begin{figure}[t] 
\centering
\begin{subfigure}[t]{.33\linewidth}
\centering
\includegraphics[width=0.95\linewidth]{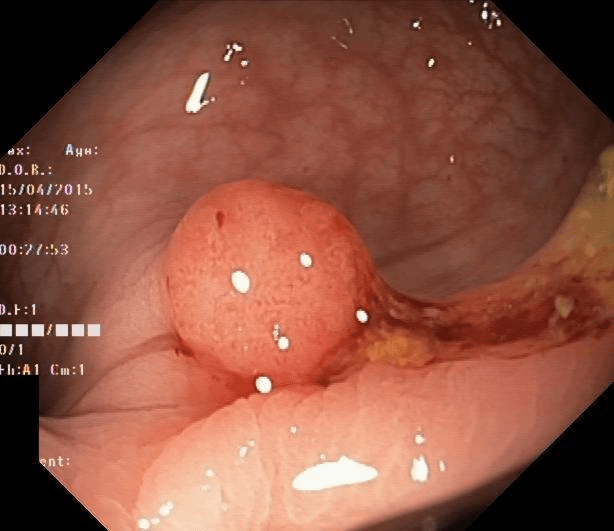}
\caption{Original image}
\label{sfig:ip_inpaint1}
\end{subfigure}%
\begin{subfigure}[t]{.33\linewidth}
\centering
\includegraphics[width=0.95\linewidth]{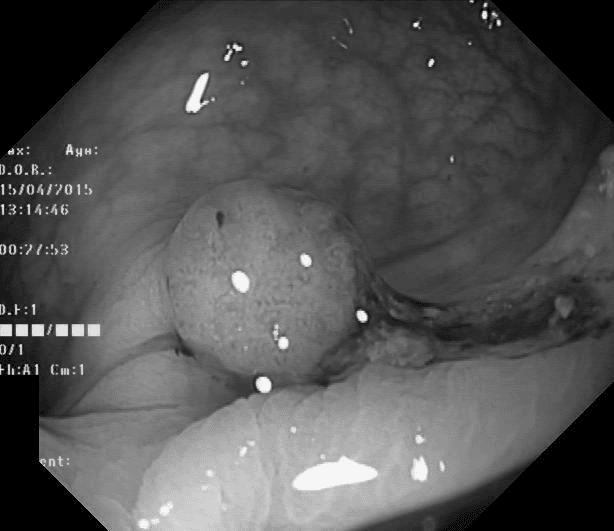}
\caption{Greyscale conversion}
\label{sfig:ip_inpaint2}
\end{subfigure}%
\begin{subfigure}[t]{.33\linewidth}
\centering
\includegraphics[width=.95\linewidth]{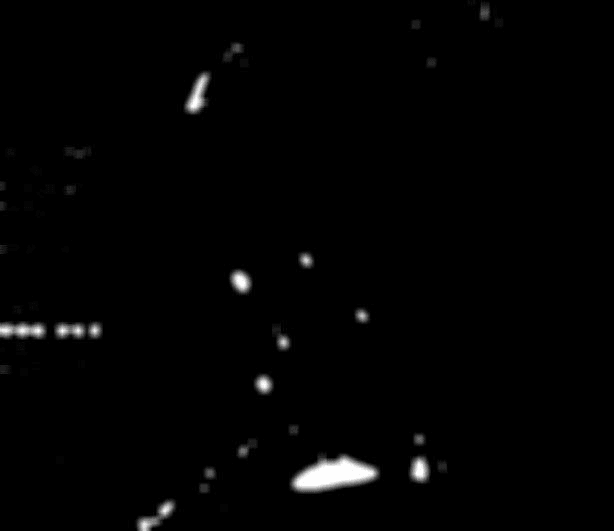}
\caption{Specular highlights detection}
\label{sfig:ip_inpaint3}
\end{subfigure}\par\medskip
\begin{subfigure}[t]{.33\linewidth}
\centering
\includegraphics[width=.95\linewidth]{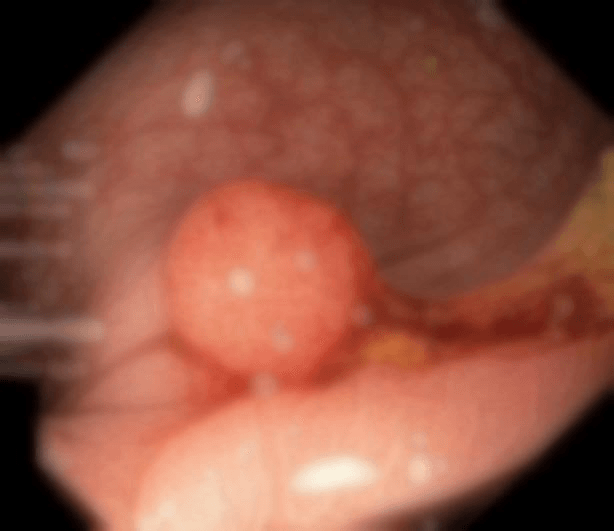}
\caption{Average smoothing}
\label{sfig:ip_inpaint4}
\end{subfigure}%
\begin{subfigure}[t]{.33\linewidth}
\centering
\includegraphics[width=.95\linewidth]{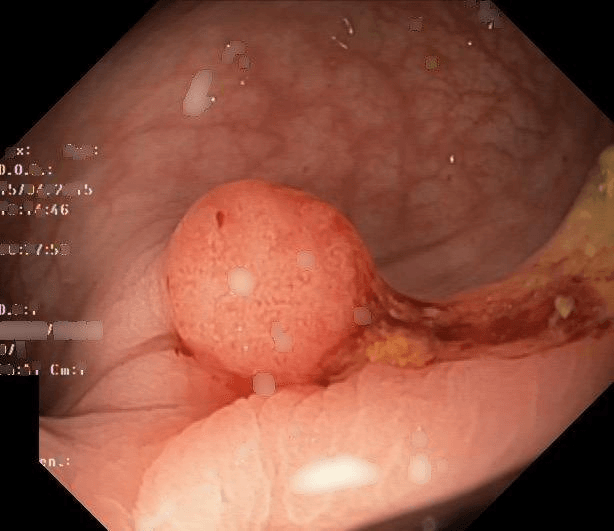}
\caption{Initial restoration}
\label{sfig:ip_inpaint5}
\end{subfigure}%
\begin{subfigure}[t]{.33\linewidth}
\centering
\includegraphics[width=.95\linewidth]{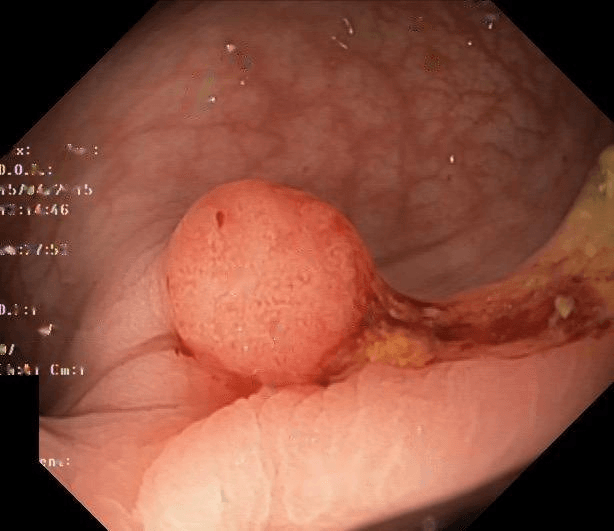}
\caption{Inpainted using Telea algorithm}
\label{sfig:ip_inpaint6}
\end{subfigure}%
\caption{An overview of stages of the specular highlights inpainting method}
\label{fig:ip_inpaint}
\end{figure}
The removal of specular highlights from colonoscopy image includes two sequential processes: detection of specular highlights and highlights inpainting.
Figure \ref{fig:ip_inpaint} depicts the overall procedure of the method, the outcome of which is generally based on the combination of Telea inpainting algorithm with initial image restoration after several modification steps.

\begin{figure}[t]
\centering
\begin{subfigure}[t]{.25\linewidth}
\centering
\includegraphics[width=0.95\linewidth]{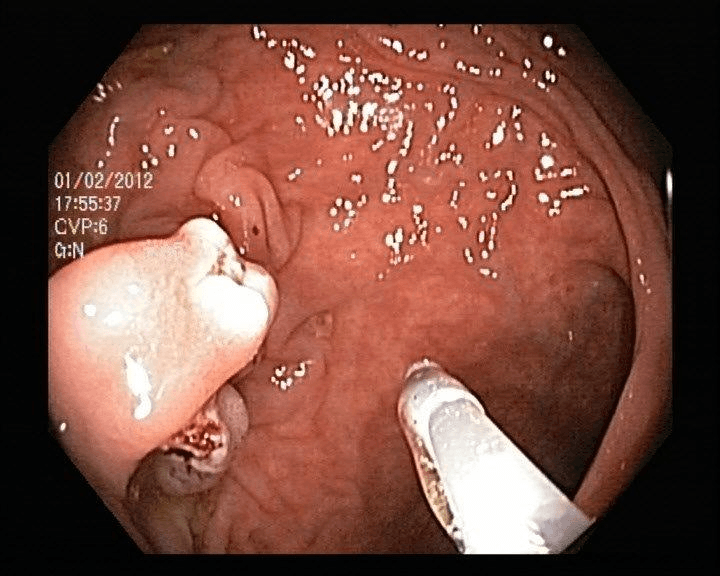}
\caption{Original image}
\label{sfig:ip_mask1}
\end{subfigure}%
\begin{subfigure}[t]{.25\linewidth}
\centering
\includegraphics[width=.95\linewidth]{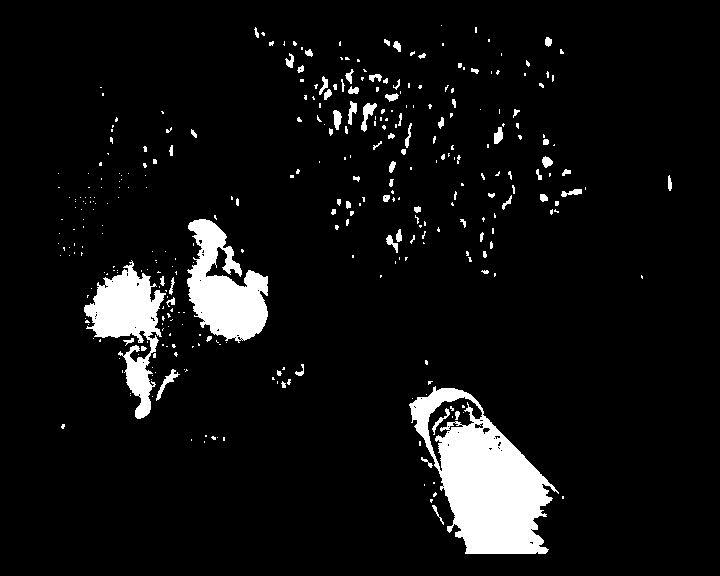}
\caption{Thresholding}
\label{sfig:ip_mask2}
\end{subfigure}
\begin{subfigure}[t]{.25\linewidth}
\centering
\includegraphics[width=.95\linewidth]{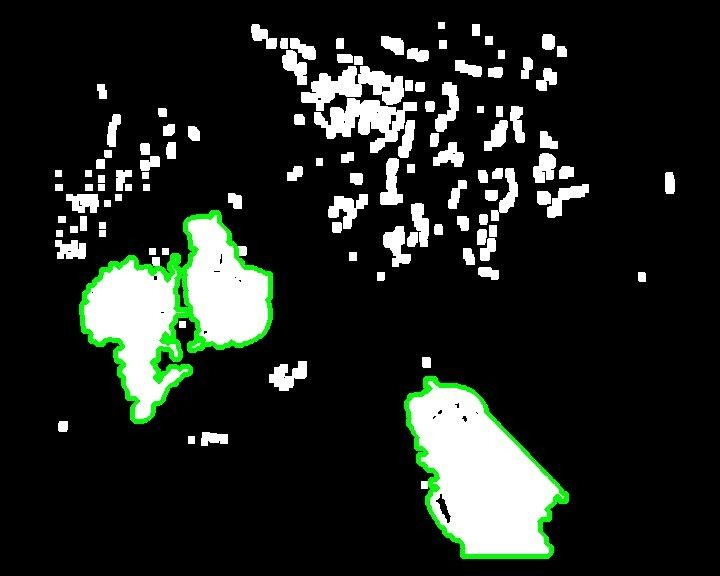}
\caption{Detection of undesired regions}
\label{sfig:ip_mask3}
\end{subfigure}%
\begin{subfigure}[t]{.25\linewidth}
\centering
\includegraphics[width=.95\linewidth]{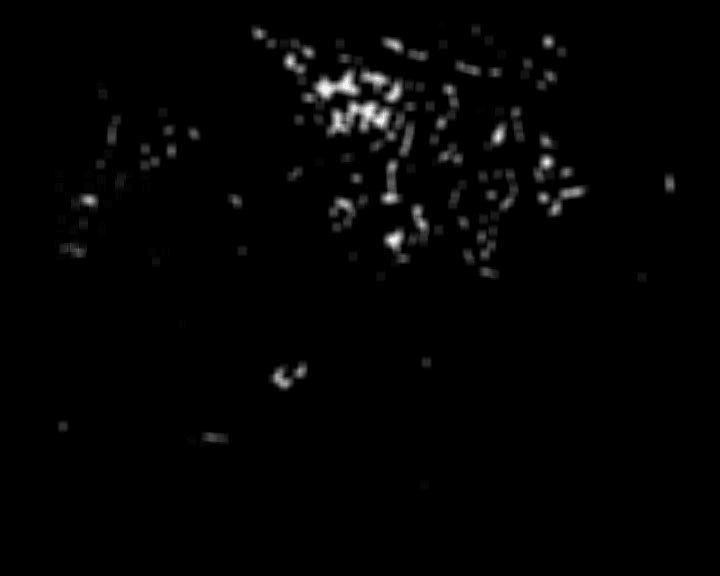}
\caption{Preprocessed mask}
\label{sfig:ip_mask4}
\end{subfigure}%
\caption{An illustration of specular highlights detection from a colonoscopy image that contains text, high exposure regions and a white instrument.}
\label{fig:ip_mask}
\end{figure}

\paragraph{Specular highlights detection} First, it is necessary to convert the image from the original RGB channel to grey scale to process the subsequent procedure. 
Rather than adaptive thresholding, the proposed approach employs standard thresholding method with a fixed threshold value to identify specular highlights in all images. This is due to the gastrointestinal image's varied textures and components, and if not done properly, may result in information loss. Some samples of the dataset contain text, high exposure regions and brightly colored instrument, as described in Figure \ref{fig:ip_mask}. Aside from text in white color, high exposure regions are parts of specular highlights that received excessively high intensity compared to regular highlight spots, while the instruments are sometimes in white or blue color. After thresholding, these factors may emerge in the mask, as shown in Figure \ref{sfig:ip_mask2}, and affect the inpainting outcome. Thus, the following step is to remove these undesired elements from the mask in order to assure consistency. To cope with these problems, two directions are considered, either to perform segmentation for text, polyp and instrument, separately, or remove the parts that meet certain size threshold. For simplicity, the second approach is used in this study.

The preprocessing step consists of several morphology transformations interspersed by contour detection and removal. More specifically, a dilation operation with kernel size $3\times3$ is performed initially to connect the pixels related to undesirable parts. Among the obtained contours, those whose scaled area following the Modified Z-scores formula \cite{HoaglinOutliers}, as shown in Formula \ref{eq1}, exceeds $17.0$ are removed from the mask. The mask is then passed into another erosion module with the same settings to restore the initial highlights intensity. Finally, Gaussian filter of size $19\times19$ is applied to reduce the intensity of highlights area and improve the inpainting performance.

\begin{equation}\label{eq1}
  S_i = \frac{|s_i-\tilde{s}|}{MAD}
\end{equation}
where:
\begin{itemize}
    \item $S_i$: is the scaled area of contour $i$ based on modified Z-score.
    \item $s_i$: is the area of contour $i$
    \item $\tilde{s}$: is the median area of all contours
    \item $MAD = median(|s_i-\tilde{s}|),\forall i=1..n$: is the Median Absolute Deviation of contour areas
\end{itemize}
\paragraph{Highlights inpainting} 
Once the mask of specular highlights has been achieved, the image regions indicated by the mask are then reconstructed through an efficient inpainting operation. First, a filter of size $3\times3$ slides across every pixels of the original image and calculate the average value. The process is repeated $N$ times to ensure a desirable outcome. We then perform an initial restoration on the image by directly replacing its pixels under the specular highlights mask with pixels from the blur image. Despite the drastically reduced intensity, specular highlight spots still remains in the reconstructed image, as shown in Figure \ref{sfig:ip_inpaint5}. To obtained the final result, Telea algorithm \cite{TeleaInpaint}, a powerful image inpainting strategy, is applied to eliminate the remaining noisy and dim highlights. The inpainted image is noticeably higher in quality, with specular highlights removed without negatively impacting other areas of the image.

\subsubsection{Black Mask Removal}

\begin{figure}[t]
\centering
\begin{subfigure}[t]{.25\linewidth}
\centering
\includegraphics[width=0.95\linewidth]{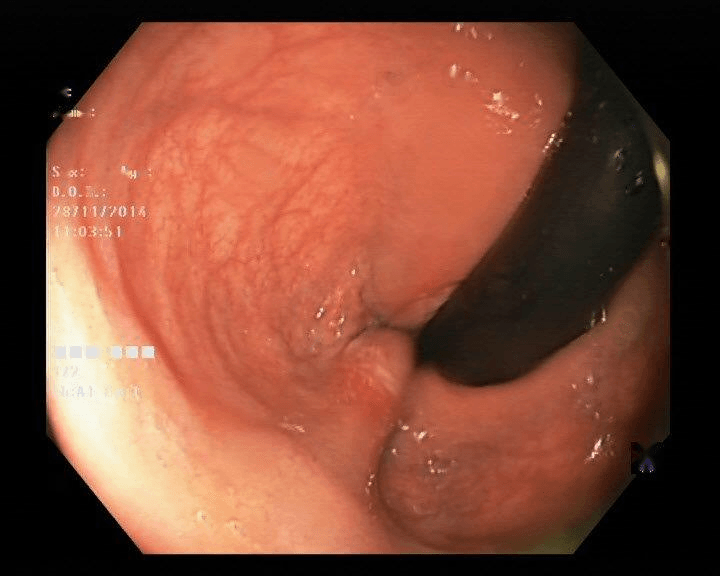}
\label{sfig:bm1}
\end{subfigure}%
\begin{subfigure}[t]{.25\linewidth}
\centering
\frame{\includegraphics[width=0.95\linewidth]{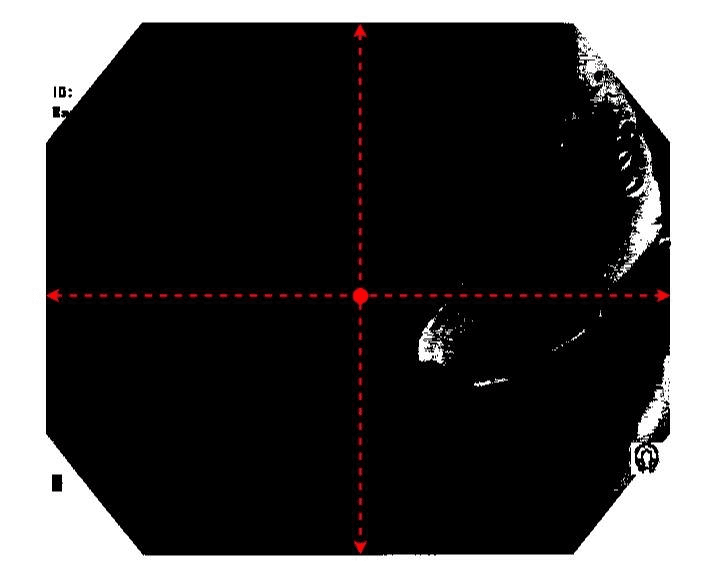}}
\label{sfig:bm2}
\end{subfigure}%
\begin{subfigure}[t]{.25\linewidth}
\centering
\frame{\includegraphics[width=.95\linewidth]{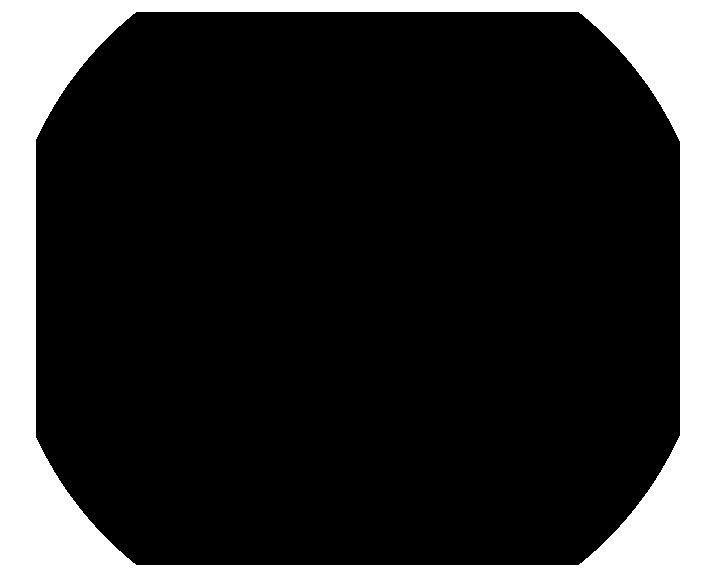}}
\label{sfig:bm5}
\end{subfigure}%
\begin{subfigure}[t]{.25\linewidth}
\centering
\frame{\includegraphics[width=.95\linewidth]{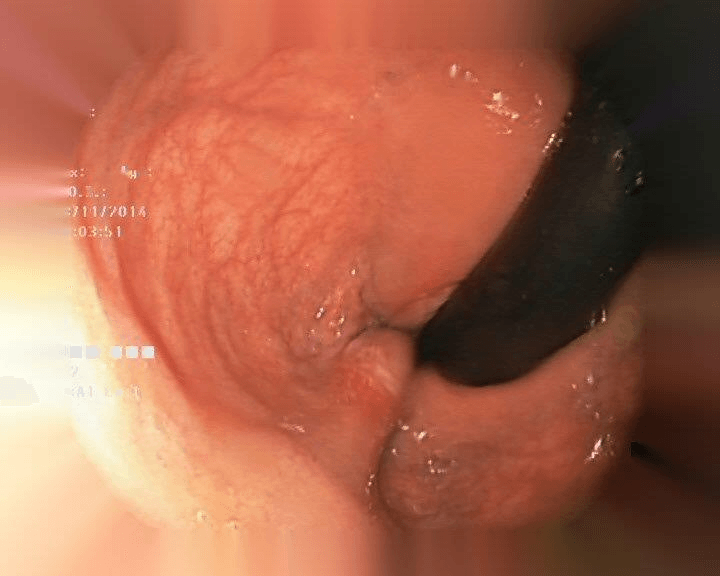}}
\label{sfig:bm6}
\end{subfigure}\par\medskip
\begin{subfigure}[t]{.25\linewidth}
\centering
\includegraphics[width=0.95\linewidth]{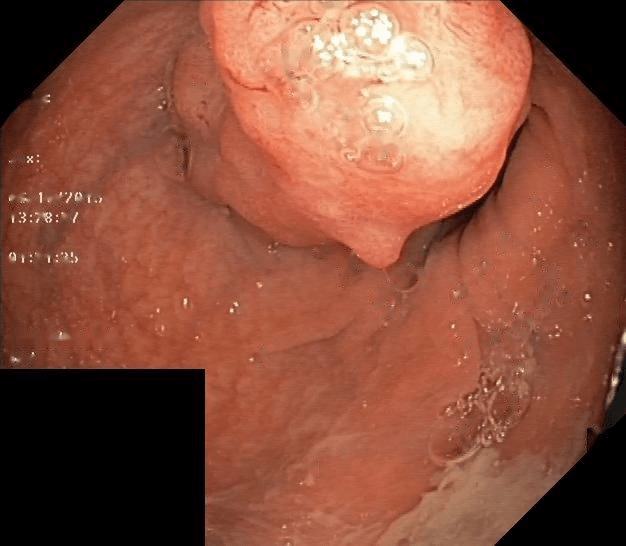}
\label{sfig:bm1b}
\end{subfigure}%
\begin{subfigure}[t]{.25\linewidth}
\centering
\frame{\includegraphics[width=0.95\linewidth]{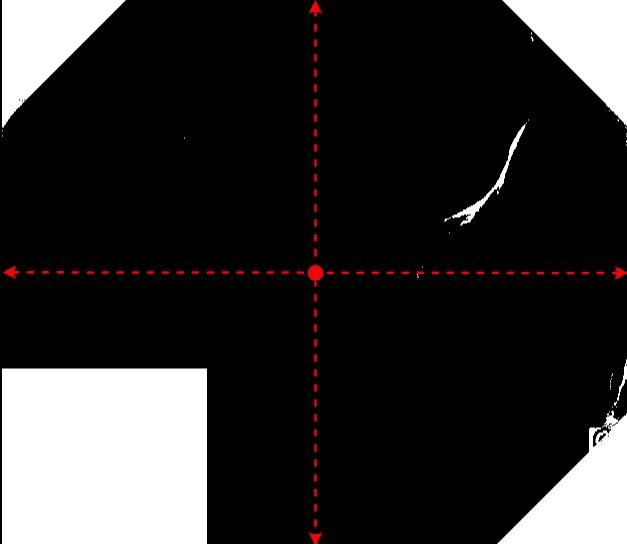}}
\label{sfig:bm2b}
\end{subfigure}%
\begin{subfigure}[t]{.25\linewidth}
\centering
\frame{\includegraphics[width=.95\linewidth]{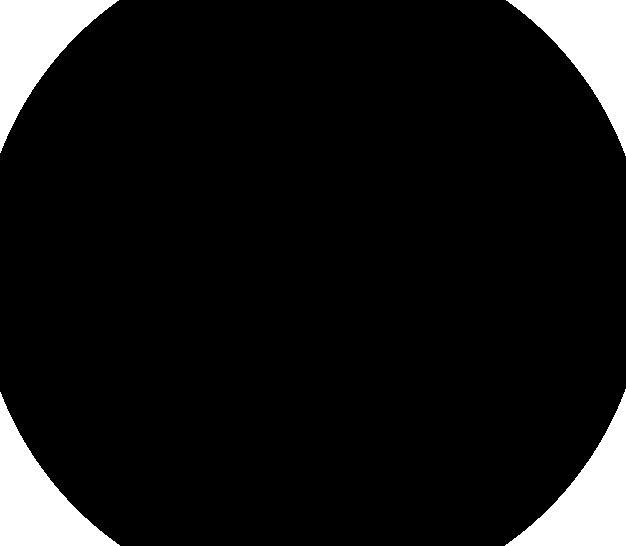}}
\label{sfig:bm5b}
\end{subfigure}%
\begin{subfigure}[t]{.25\linewidth}
\centering
\frame{\includegraphics[width=.95\linewidth]{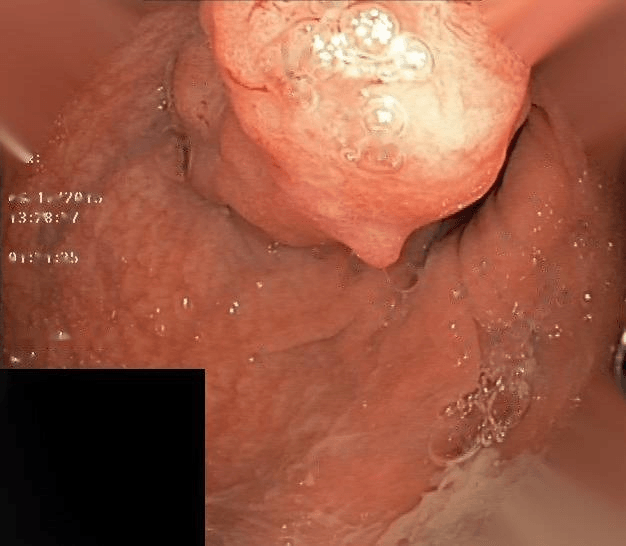}}
\label{sfig:bm6b}
\end{subfigure}\par\medskip
\begin{subfigure}[t]{.25\linewidth}
\centering
\includegraphics[width=0.95\linewidth]{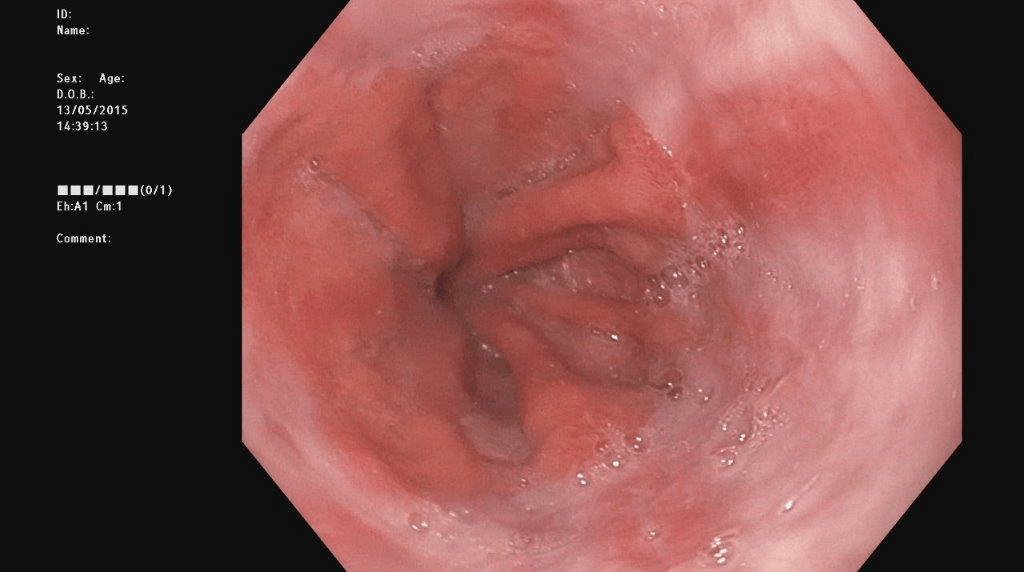}
\caption{Image with black mask}
\label{sfig:bm1c}
\end{subfigure}%
\begin{subfigure}[t]{.25\linewidth}
\centering
\frame{\includegraphics[width=0.95\linewidth]{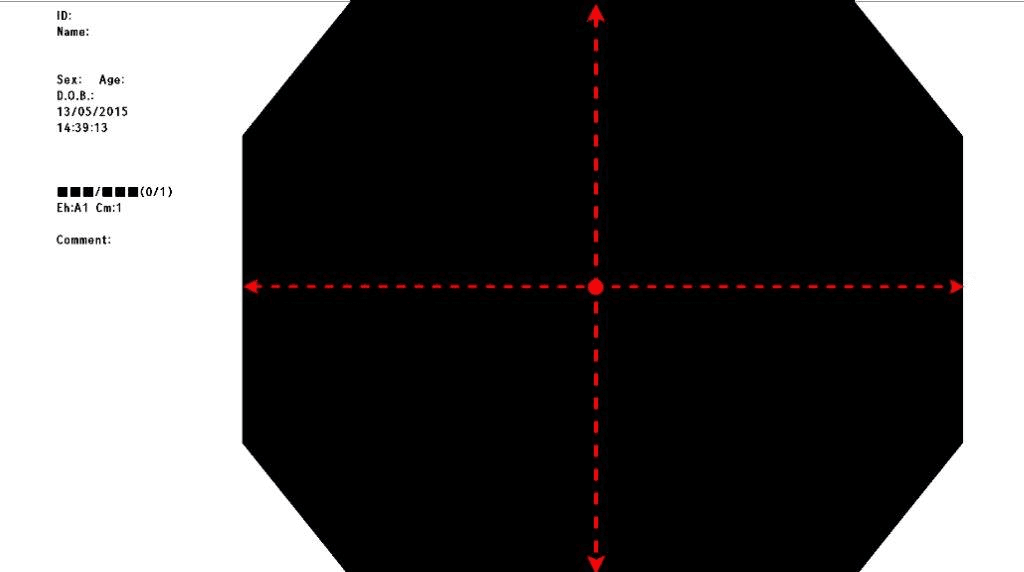}}
\caption{Border detection}
\label{sfig:bm2c}
\end{subfigure}%
\begin{subfigure}[t]{.25\linewidth}
\centering
\frame{\includegraphics[width=.95\linewidth]{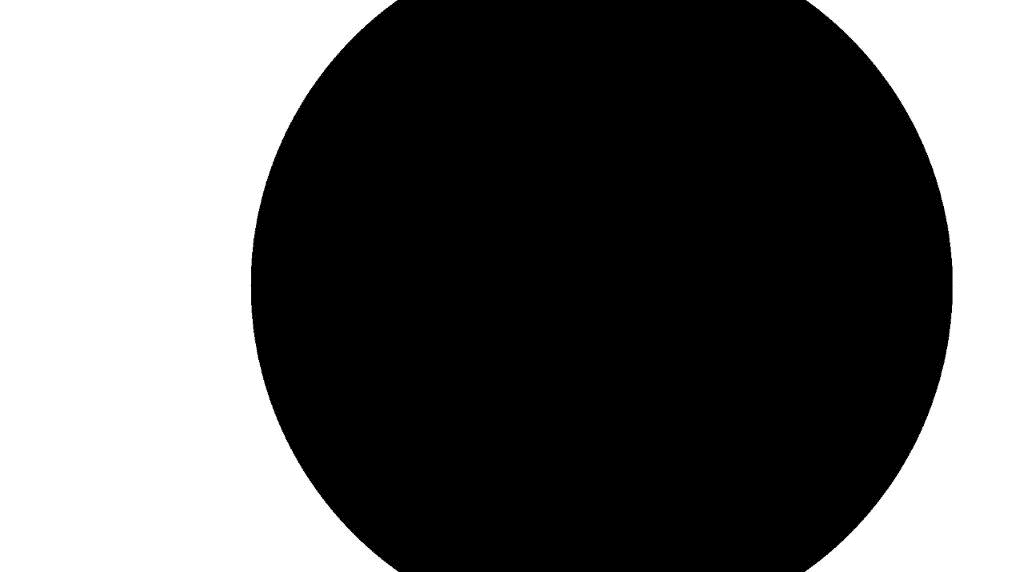}}
\caption{Artificial mask}
\label{sfig:bm5c}
\end{subfigure}%
\begin{subfigure}[t]{.25\linewidth}
\centering
\frame{\includegraphics[width=.95\linewidth]{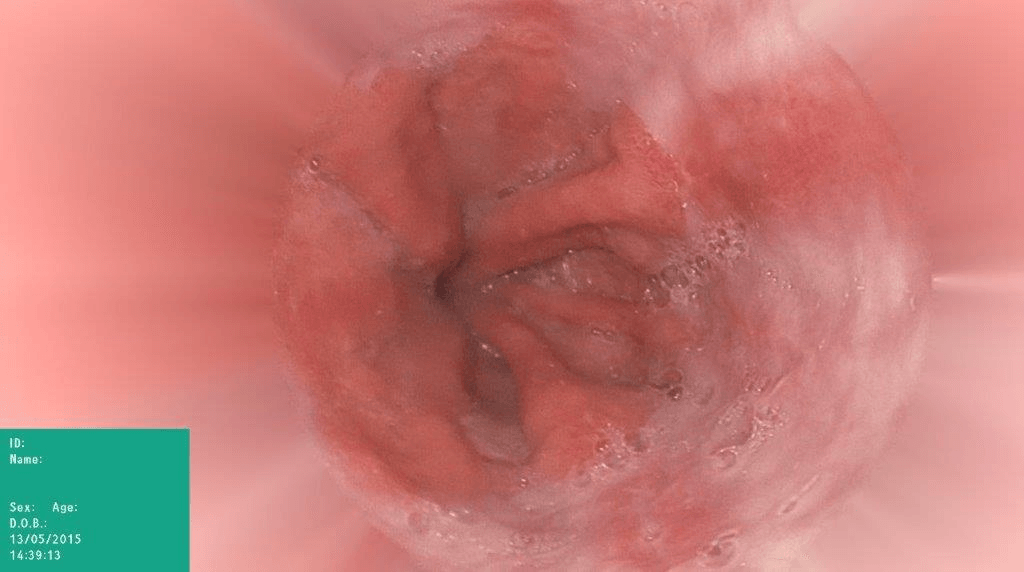}}
\caption{Enhanced Image}
\label{sfig:bm6c}
\end{subfigure}%
\caption{Stages of black mask removal process. The first row illustrates an image with a standard black mask, the second row depicts an image containing a black square and the last row contains image with black mask marked as black box artefact.}
\label{fig:bm}
\end{figure}
Previous research has shown that black masks do generate valley information, which can reduce polyp localization performance. Based on this, we propose a black mask removal strategy for the VQA task that still retains black box information in order to answer the question "Is there a green/black box artefact?". In general, an artificial mask of black frame is initially created based on its border width, and then the inpainting operation is performed to remove the black frame from the image. The overall procedure is described in Figure \ref{fig:bm}. Our method does not use cropping or thresholding directly to detect and remove the black mask because it may contain the black box artifact, shadow regions, or black instrument, the removal of which causes information loss and decreases VQA performance.

To detect the border width, we first perform a grey scale conversion and inverse thresholding with erosion operation to remove noise, and then measure the distance from each edge of the image to the nearest pixel that does not belong to the mask. After determining the width of the border, the crucial step of the method is to create an artificial mask with internal octagon shape. This can be done by creating two sub-masks, one rectangle and one circle, followed by a bitwise OR operation to combine them into the final mask, as show in Figure \ref{sfig:bm5c}. The circle mask is created with a center point based on the information of border width and a radius calculated by multiplying the ordinate of the center point by a value $\sigma$ $(\sigma>1)$. In some cases, the final mask is not always octagonal, as shown in the last example, but it still covers the main region of interest. Finally, the inpainting of black mask is completed using the same procedure as described in the previous section for specular highlights, giving the final enhanced image with black mask removed. If a black box artefact exists in the bottom-left corner, as shown in the second example, it will not be significantly affected as long as its size is greater than the area of the mask at the respective position. For images containing an expanded black mask labeled as black box artefact, we process by creating a simulated green box that contains the text and placing it in bottom-left corner. By doing so, the text and box artefact information still remain after the inpainting procedure. Though the obtained results are quite satisfactory, there are still some cases where the mask is not completely removed and need further processing steps.

\subsection{Multimodal Fusion Architecture}
Since this study focus mainly on the VQA task, the architecture should be capable of extracting meaningful features from the question and corresponding image, and incorporating them to give the correct answer. Our multimodal fusion architecture is set up with important components such as an image encoder for feature extraction from images, a text encoder for features extraction from questions, a fusion algorithm for unifying modalities and a classifier for producing the appropriate answer. The proposed approach uses pre-trained Bidirectional Encoder Representations based on Transformers (BERT) \cite{kenton2019bert} to extract textual features from questions. As a bidirectional model, it can learn the meaning of words in a sentence by considering both the words that come before and after them. With massive pre-training data, BERT can be fine-tuned and achieved state-of-the-art results on a number of natural language processing (NLP) benchmarks. For features extraction from the images, this study set up and experiment with eight different pre-trained models that are belong to two main concepts:
\begin{itemize}
    \item CNN-based architectures including \textbf{Resnet152} \cite{7780459},  \textbf{Inception-v4} \cite{10.5555/3298023.3298188}, \textbf{MobileNetV2} \cite{8578572}
     and \textbf{EfficientNet} \cite{tan2020efficientnet}. The group of models take advantage of traditional CNN's components such the convolutional layer, pooling layer, residual block and fully connected layer to achieve significant result in computer vision field. The training of CNN-based model is more efficient
     with less computational resources compared to new approaches based on Transformers.

    \item Transformer-based architectures including \textbf{ViT} \cite{dosovitskiy2020vit}, 
    \textbf{DeiT} \cite{pmlr-v139-touvron21a}, \textbf{Swin Transformer} \cite{liu2021swin} and \textbf{BEiT} \cite{bao2022beit}. The family of models leverages a massive amount of training data and Transformer's multi-head self-attention for a game-changing breakthrough in the computer vision field. ViT (Vision Transformer) and other models inspired from it initially encodes the image as patch embeddings and pass them into a regular  Transformer Encoder for feature extraction, which is similar to text data.
    Currently they are considered as the prominent architectures to achieve state-of-the-art performance on a variety of tasks in computer vision such as image classification, object detection, and semantic image segmentation.
    
\end{itemize}

After obtaining the embeddings of text and image, a multimodal fusion method based on concatenation is used to combine
 these features along the embedding dimension. The unified embedding matrix is then passed through an intermediate fully connected layer with drop out 0.5 and ReLU activation followed by a classification layer to produce the final output. Because there can be more than one appropriate answer for each question, we approach the VQA task as a multi-label classification problem. To successfully train the proposed arhitecture, multi-label binarization is used to encode a list of all possible answers into a binary vector. Furthermore, the final layer is configured with sigmoid activation function to return an output vector of the same size containing the corresponding probability for each class.
\section{Experimental Setup}

\subsection{Data Preparation}
The development set released for the VQA challenge contains 2000 images of  gastroscopy and colonoscopy procedures. In order to experiment and evaluate our method, we randomly divided the provided development set into three parts: train, validation, and test, with 1600 images for training and 200 images for each validation and test set. The data preparation process is designed to ensure that each abnormality has the same proportion in the training, validation, and testing sets, and that each image contains all 18 questions. This produces 28,800 question-answer pairs on the training set, 3600 pairs for validation and 3600 pairs for test.

All images from development set and private test set are first passed into an image enhancement block, where numerous image preprocessing methods are applied to remove specular highlights and black mask from the images. The enhanced results are then used as input in the training and testing of the proposed VQA model.

\subsection{Experiment Configurations}
\begin{table*}
  \caption{Statistics of multimodal fusion with pre-trained vision and language models for the VQA challenge}
  \label{tab:pre-trained}
  \resizebox{\textwidth}{!}{%
  \begin{tabular}{lclc}
    \toprule
    \textbf{Models} & \textbf{Version} & \textbf{Spaces vision model name} &\textbf{\# Parameters}\\
    \midrule
    BERT+ViT & base & "google/vit-base-patch16-224-in21k" & 196M\\
    BERT+DEiT & base & "facebook/deit-base-distilled-patch16-224" & 196M\\
    BERT+Swin & base & "microsoft/swin-base-patch4-window7-224-in22k" & 197M\\
    BERT+ BEiT& base & "microsoft/beit-base-patch16-224-pt22k-ft22k" & 196M\\
    BERT+ResNet152 & v1.5 & "microsoft/resnet-152" & 169M\\
    BERT+Inception & v4 & "inception\_v4" & 153M\\
    BERT+MobileNet & V2 & "google/mobilenet\_v2\_1.0\_224" & 112M\\
    BERT+EfficientNet & b3 & "google/efficientnet-b3" & 121M\\
  \bottomrule
\end{tabular}}
\end{table*}

Many experiments are carried out in order to evaluate the performance of the proposed methods toward the ImageCLEFmed-MEDVQA-GI-2023 challenge. Specifically, each pre-trained vision model is initialized and experimented as an image encoder and unify with BERT encoder through concatenation fusion for multimodal learning. Table \ref{tab:pre-trained} gives the general information of pre-trained models used in this study including vision model name, version and number of parameters for each fusion model. Through experiments, we can discover the potential and limitation of each model for the VQA task and thus, choose the best method for giving the final prediction on the private test set of the competition.

To achieve a comparative result, we set up the same hyperparameters for all experiments. The models are trained in 15 epochs with batch size of 64. We utilize the Adam optimizer \cite{kingma2014adam} using weighted decay with an initial learning rate of 5e-5 and a linear scheduler to decrease learning rate 6.67\% after each epoch. Since we approach the VQA task as multi-label classification, the output layer is configured to return a tensor containing probabilities of answers, where the final predicted answers for each question can be achieved using threshold value of 0.5. Due to this, the BCEWithLogitsLoss function, which combines a Sigmoid layer and the BCELoss, is applied in the training process. After each epoch, the training loss and validation loss are calculated, and the performance are then evaluated on classification metrics such as accuracy, precision, recall and F1-Score. To ensure a meaningful result for multi-label classification, the metrics are calculated using ground truth and prediction sets of binary vectors, in which recall, precision and F1-scores should be calculated on each sample and find their average. The model's state that obtains best F1-Score is used for prediction in the testing phase.

The proposed architecture are implemented in PyTorch and trained on the Kaggle platform with hardware specifications: Intel(R) Xeon(R) CPU @ 2.00GHz; GPU Tesla P100 16 GB with CUDA 11.4.
\section{Experimental Results}
\begin{table*}[!htbp]
\caption{Comparative performance of the multimodal fusion method with vision models on the development test set.}
\label{tab:result}
\renewcommand{\arraystretch}{1.05}
\centering
\begin{tabular}{p{4cm}cccc}\toprule
                         \textbf{Vision Models} & \multicolumn{1}{c}{\textbf{Accuracy}} & \multicolumn{1}{c}{\textbf{Precision}} & \multicolumn{1}{c}{\textbf{Recall}} & \multicolumn{1}{c}{\textbf{F1-Score}} \\\midrule
\multicolumn{5}{l}{\textbf{No image enhancement}\vspace{0.3em}}                                                                             \\
ResNet152       & 0.8419                                & 0.8917                                 & 0.8867                              & 0.8857
\\
Inception-v4     & 0.8619                                & 0.9133                                 & 0.9067                              & 0.9067                          \\
MobileNetV2     & 0.8444                                & 0.8932                                 & 0.8951                              & 0.8906                          \\
EfficientNet-B3 & 0.8581                                & 0.9065                                 & 0.9049                              & 0.9023\vspace{0.3em}                                                             \\
ViT-B/16        & 0.8636                                & 0.9134                                 & 0.9089                              & 0.9078                          \\
DeiT-B          & 0.8611                                & 0.9100                                 & 0.9026                              & 0.9033                          \\
Swin-B          & \textbf{0.8664}                       & 0.9152                                 & \textbf{0.9094}                     & \textbf{0.9090}                 \\
BEiT-B          & 0.8647                                & \textbf{0.9158}                        & 0.9068                              & 0.9074                                                                             \\\midrule
\multicolumn{5}{l}{\textbf{With image enhancement}\vspace{0.3em}}                                                                                                                                                           \\
ResNet152       & 0.8453 $\small\uparrow$                               & 0.8942                                 & 0.8894                              & 0.8885 $\small\uparrow$ \\
Inception-v4     & 0.8625 $\small\uparrow$                                & 0.9121                                 & 0.9073                              & 0.9071 $\small\uparrow$                          \\
MobileNetV2     & 0.8422 $\small\downarrow$                                & 0.8935                                 & 0.8882                              & 0.8867 $\small\downarrow$                          \\
EfficientNet-B3 & 0.8572 $\small\downarrow$                                & 0.9081                                 & 0.9079                              & 0.9046 $\small\uparrow$\vspace{0.3em}\\
ViT-B/16        & 0.8631 $\small\downarrow$                                & 0.9126                                 & 0.9086                              & 0.9073 $\small\downarrow$                          \\
DeiT-B          & 0.8625 $\small\uparrow$                                & 0.9122                                 & 0.9052                              & 0.9055 $\small\uparrow$                          \\
Swin-B          & 0.8717 $\small\uparrow$                                & 0.9245                                 & 0.9159                              & 0.9168 $\small\uparrow$                         \\
BEiT-B          & \textbf{0.8725 $\small\uparrow$}                        & \textbf{0.9253}                        & \textbf{0.9184}                     & \textbf{0.9185 $\small\uparrow$}                                                    
\\\bottomrule
\end{tabular}
\end{table*}
\vspace{5em}
The comparative result of different pre-trained image model on the testing set is shown in Table~\ref{tab:result}. It is clear that, with no image enhancement, Swin-B achieves the best result with 86.64\% accuracy and  90.90\% F1-Score while BEiT-B gives a slightly lower performance with accuracy of 86.47\% and 90.74\% F1-Score. CNN-based vision models have acceptable results, but cannot be compared with the result of Transformer architecture models.


With image enhancement,
six out of eight vision models from both CNN and Transformer architectures achieve a better performance on F1-Score metric. BEiT-B has an outstanding result with accuracy and F1-Score of 87.2\% and 91.85\%, respectively. Overall, the enhancement process helps to improve the F1-Score at least 0.4\% and up to 1.11\% on VQA performance. The result of the convolutional models is still under when compared with Transformers architecture models.

 We found that the BERT and BEiT fusion (BERT+BEiT) with image enhancement is the best method of our approach and use it for prediction in final private test phase. Our method obtains a good result on the private test set with an accuracy of 82.01\%. Table \ref{tab:final} illustrates the performance evaluation of BERT+BEiT fusion on each question from the development test set compared with the private test set. In general, there are 14/18 questions with predicted answers achieve greater than 80\% accuracy on the development test set, while 11/18 questions on the private test set achieve the same result. Our method still struggles to produce full and precise answers for questions with multiple answers, such as "What color is the abnormality?" or questions that refer to the location of the abnormality, anatomical landmark, and instrument.

\begin{table}[]
\centering
\caption{Performance evaluation of BERT+BEiT fusion with image enhancement for each question on the development test set and private test set}
\label{tab:final}
\setlength{\tabcolsep}{0.3em}
\resizebox{\textwidth}{!}{%
\begin{tabular}{lccccc}
\toprule
\multirow{2}{*}{\textbf{Question}}                        & \multicolumn{4}{c}{\textbf{Development Test Set}}                                      & \textbf{Private Test Set} \\
                                                          & \textbf{Accuracy } & \textbf{Precision} & \textbf{Recall} & \textbf{F1-Score}     & \textbf{Accuracy}         \\\midrule
\textbf{Are there any abnormalities in the image?}        & 0.9700            & 0.9750             & 0.9725          & 0.9733          & 0.8091                    \\
\textbf{Are there any anatomical landmarks in the image?} & 0.9300            & 0.9300             & 0.9300          & 0.9300          & 0.6940                    \\
\textbf{Are there any instruments in the image?}          & 0.9050            & 0.9275             & 0.9200          & 0.9217          & 0.7688                    \\
\textbf{Have all polyps been removed?}                    & 0.9550            & 0.9575             & 0.9600          & 0.9583          & 0.9721                    \\
\textbf{How many findings are present?}                   & 0.8400            & 0.8400             & 0.8400          & 0.8400          & 0.7807                    \\
\textbf{How many instrumnets are in the image?}           & 0.9650            & 0.9650             & 0.9650          & 0.9650          & 0.8901                    \\
\textbf{How many polyps are in the image?}                & 0.9650            & 0.9650             & 0.9650          & 0.9650          & 0.9577                    \\
\textbf{Is there a green/black box artefact?}             & 0.9500            & 0.9500             & 0.9500          & 0.9500          & 0.9732                    \\
\textbf{Is there text?}                                   & 0.9250            & 0.9250             & 0.9250          & 0.9250          & 0.8787                    \\
\textbf{Is this finding easy to detect?}                  & 0.8900            & 0.8900             & 0.8900          & 0.8900          & 0.8044                    \\
\textbf{What color is the abnormality?}                   & 0.5800            & 0.9025             & 0.8563          & 0.8597          & 0.4969                   \\
\textbf{What color is the anatomical landmark?}           & 0.9400            & 0.9400             & 0.9400          & 0.9400          & 1.0000                    \\
\textbf{What is the size of the polyp?}                   & 0.8600            & 0.8650             & 0.8700          & 0.8667          & 0.8535                    \\
\textbf{What type of polyp is present?}                   & 0.8650            & 0.8800             & 0.8725          & 0.8750          & 0.8132                    \\
\textbf{What type of procedure is the image taken from?}  & 1.0000            & 1.0000             & 1.0000          & 1.0000          & 0.9938                    \\
\textbf{Where in the image is the abnormality?}           & 0.6600            & 0.9251             & 0.8805          & 0.8842          & 0.5872                   \\
\textbf{Where in the image is the anatomical landmark?}   & 0.7150            & 0.8847             & 0.8848          & 0.8766          & 0.7203                    \\
\textbf{Where in the image is the instrument?}            & 0.7900            & 0.9332             & 0.9096          & 0.9125          & 0.7688                    \\\midrule
\textbf{All}                                              & \textbf{0.8725}   & \textbf{0.9253}    & \textbf{0.9184} & \textbf{0.9185} & \textbf{\underline{0.8201}} \\\bottomrule         
\end{tabular}}
\end{table}

\section{Conclusion and Future Works}
Along with performing image enhancement, we also set up and experimented using various powerful pre-trained image models together with the BERT encoder for our proposed multimodal architecture  in the VQA task at ImageCLEFmed-MEDVQA-GI-2023 \cite{ImageCLEFmedicalCaptionOverview2023}. The visual enhancement steps, which include specular hightlights and black mask removals, help improve multimodal learning performance on the dataset by up to 1.11\% F1-Score. Our best method, BERT+BEiT fusion with image enhancement, achieved 87.25\% on development test set and 82.01\% on the private test set by the accuracy. Through performance analysis, there are question cases that require multiple positions or colors in the answer, which are our limitations in this study. In summary, there are factors that have significant impact on our solution for the VQA task such as answer imbalance, noise, and artifacts.

Our future research for this task is to improve the accuracy of the model in giving the correct answer by enriching the features from images and questions through instrument segmentation and polyp localization with methods such as U-net \cite{DBLP:journals/corr/RonnebergerFB15}, ResUnet++ \cite{8959021} developed on object-specific datasets such as Kvasir-Instrument \cite{10.1007/978-3-030-67835-7_19} and Kvasir-seg \cite{10.1007/978-3-030-37734-2_37}. Other advanced colonoscopy image preprocessing techniques such as interlacing removal or uneven lighting removal can be examined to improve the image quality. From the proposed system, an intelligent chatbot application can be implemented for question-answering from medical images and help improve colonoscopy analysis.

\begin{acknowledgments}
This research was supported by The VNUHCM University of Information Technology’s Scientific Research Support Fund.
\end{acknowledgments}

\bibliography{main}




\end{document}